\documentclass[conference]{IEEEtran}
\IEEEoverridecommandlockouts
\usepackage{cite}
\usepackage{amsmath,amssymb,amsfonts}
\usepackage{algorithmic}
\usepackage{graphicx}
\usepackage{textcomp}
\usepackage{xcolor}
\usepackage{mathtools}
\usepackage{xspace}
\usepackage{siunitx}

\usepackage[a4paper, total={184mm,239mm}]{geometry}
\def\BibTeX{{\rm B\kern-.05em{\sc i\kern-.025em b}\kern-.08em
    T\kern-.1667em\lower.7ex\hbox{E}\kern-.125emX}}

\newcommand{\ip}[2]{\langle #1, #2 \rangle}
\newcommand{\lTwoSq}[1]{\left\lVert #1 \right\rVert_2^2}
\newcommand{\lTwo}[1]{\left\lVert #1 \right\rVert_2}

\newcommand{\design}{FaTRQ\xspace}

\IEEEaftertitletext{\vspace{-1.5\baselineskip}}

\begin{document}

\title{FaTRQ: Tiered Residual Quantization for LLM Vector Search in Far-Memory-Aware ANNS Systems}

\author{\IEEEauthorblockN{Tianqi Zhang}
\IEEEauthorblockA{\
\textit{UC San Diego}\\
La Jolla, CA, USA \\
tiz014@ucsd.edu}
\and
\IEEEauthorblockN{Flavio Ponzina}
\IEEEauthorblockA{
\textit{San Diego State University}\\
San Diego, CA, USA \\
fponzina@sdsu.edu}
\and
\IEEEauthorblockN{Tajana Rosing}
\IEEEauthorblockA{
\textit{UC San Diego}\\
La Jolla, CA, USA \\
tajana@ucsd.edu}
}


\maketitle

\begin{abstract}
Approximate Nearest-Neighbor Search (ANNS) is a key technique in retrieval-augmented generation (RAG), enabling rapid identification of the most relevant high-dimensional embeddings from massive vector databases. Modern ANNS engines accelerate this process using prebuilt indexes and store compressed vector-quantized representations in fast memory. However, they still rely on a costly second-pass refinement stage that reads full-precision vectors from slower storage like SSDs. For modern text and multimodal embeddings, these reads now dominate the latency of the entire query. 
We propose \design, a far-memory-aware refinement system using tiered memory that eliminates the need to fetch full vectors from storage. It introduces a progressive distance estimator that refines coarse scores using compact residuals streamed from far memory. Refinement stops early once a candidate is provably outside the top-k. To support this, we propose tiered residual quantization, which encodes residuals as ternary values stored efficiently in far memory. A custom accelerator is deployed in a CXL Type-2 device to perform low-latency refinement locally. Together, \design improves the storage efficiency by 2.4$\times$ and improves the throughput by up to 9$ \times$ than SOTA GPU ANNS system.
\end{abstract}


\section{Introduction}

Nearest‑neighbor search (NNS) is the key component of modern data systems supporting tasks like semantic retrieval, recommendations, fraud detection, and large-scale ranking. While exact NNS becomes intractable at scale, approximate nearest-neighbor search (ANNS) offers high recall at much lower cost, and is now standard in production-scale systems.

\begin{figure}[tp]
    \centering
    \includegraphics[width=0.95\linewidth]{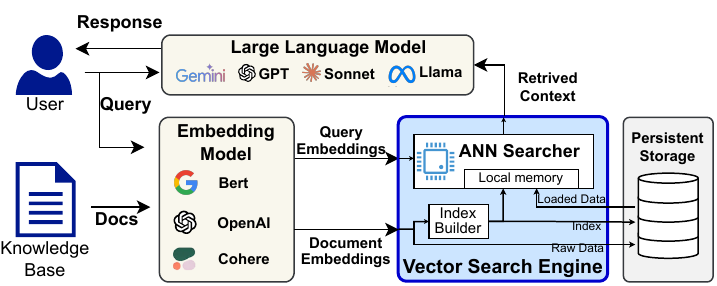}
    \vspace{-6pt}
    \caption{ANNS is the key component of the RAG pipeline.}
    \vspace{-18pt}
    \label{fig:intro}
\end{figure}

A leading use case of ANNS is retrieval-augmented generation (RAG). As shown in Figure~\ref{fig:intro}, RAG pipelines embed each document chunk once, cache the resulting vectors, and, at query time, embed the user prompt and launch an ANNS search whose results are passed to the language model. Modern dense embedding models such as OpenAI’s models~\cite{openai} generate 1536-dimensional vectors, corresponding to about 6~kB per vector in full precision. For knowledge bases containing millions or even billions of entries~\cite{milvus}, this quickly exceeds the capacity of main memory and renders exact search infeasible.

To reduce memory pressure and accelerate search,  modern engines combine indexing data structures (e.g., IVF~\cite{faiss}) with compact vector codes, such as the product quantization family~\cite{opq,dpq,lpq}. These methods quantize high-dimensional vectors into short codes, shrinking, for instance, a 6 kB floating-point vector to about 200B, so that the full index can fit in main memory. At query time, the graph guides the search process, and distances are estimated with the compact codes.

However, quantization comes at a cost: to recover full accuracy, systems re-rank the long candidate list by fetching original full-precision vectors from storage and recomputing exact distances. For modern dense embeddings~\cite{openai,sbert}, this refinement step now dominates query latency. Our profiling detailed in Section~\ref{sec:bg_anns} shows that over 90\% of query time can be spent on reading vectors from storage. Worse, the widening gap between memory and storage bandwidth makes this bottleneck increasingly severe.

At the same time, the tiered memory technologies such as CXL-based memory expander and storage-class memory (SCM) have opened opportunities to rethink this refinement stage.
While compact codes are much smaller than full-precision vectors, the aggregate working set of these codes can still exceed local DRAM (fast memory) capacity, especially for large-scale databases or multi-tenant workloads. In such cases, these tiered memories (far memory) offer an attractive option: they sit between local DRAM and SSD in both capacity and latency, making them well-suited for storing compact intermediate data structures rather than full vectors. Yet, existing ANNS systems~\cite{cxlanns,cuvs,hmann,diskann} rarely exploit these tiers and continue to treat refinement I/O as an unavoidable bottleneck.

To leverage this opportunity, we propose \textbf{\design}, a \textbf{\underline{F}}ar-memory-\textbf{\underline{a}}ware refinement system with \textbf{\underline{T}}iered \textbf{\underline{R}}esidual \textbf{\underline{Q}}uantization that eliminates most expensive I/O in high-accuracy ANNS. 
The key idea is progressive distance estimation: rather than fetching full-precision vectors, \design incrementally refines coarse distances using compact residual codes stored in far memory.  To support this, \design introduces \textit{tiered residual quantization} (TRQ), where the “T” reflects both its \textit{tier-aware} memory design and its \textit{ternary} codebook. TRQ encodes the residual between a record and its coarse quantization into ternary values, stored in far memory and streamed during query time. The estimator builds on the vector decomposition and a light-weight distance estimation calibration model, allowing residual contributions to be accumulated without reconstructing full vectors. Operating directly on ternary codes, it refines distances using only additions and subtractions, avoiding multiplications while preserving high accuracy. Though \design is applicable in software on tiered memory systems, we also built a proof-of-concept CXL Type-2 accelerator with lightweight custom logic, showing how \design can be integrated into emerging far-memory hardware.

We summarize our contributions as follows:
\begin{itemize}
\item Multi-level vector quantization for tiered memory: We design a hierarchical compression format where coarse quantized vectors reside in fast memory, and compact residual codes are stored in slower memory tiers, enabling fine-grained accuracy control with minimal I/O.
\item Progressive distance estimation: We propose an incremental scoring mechanism that updates coarse distances without reconstructing vectors, reducing memory traffic from hundreds of bytes to 4 bytes.

\item Accelerated refinement pipeline: We prototype a far-memory accelerator for residual refinements, reducing data movement and host CPU overhead.
 
\item Key Results: Improves the ANNS throughput by $2.6\times$ to $9.4\times$ compared to GPU ANNS system~\cite{cuvs}, and enhances storage efficiency by 2.4$\times$ than the refinement scheme in the SoTA pipeline~\cite{bang}.
\end{itemize}

\section{Background and Related Works}
\subsection{ ANNS System}
\label{sec:bg_anns}

\begin{figure}
    \centering
    \includegraphics[width=0.9\linewidth]{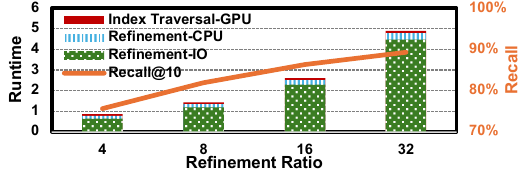}
    \vspace{-12pt}
    \caption{Runtime breakdown of IVF-refinement ANNS system}
    \vspace{-16pt}
    \label{fig:bg_prof}
\end{figure}

Modern ANNS engines combine indexes (e.g., IVF~\cite{faiss}, CAGRA~\cite{cagra}) with vector quantization to keep large-scale collections in memory while supporting fast distance estimation~\cite{fusionann, diskann,faiss,rummy,bang, cagra}. The index prunes the search space, while quantized codes enable low-cost scoring. During index traversal, the system fetches quantized codes and identifies a subset of candidates. In RAG workloads, however, modern high-dimensional embeddings (e.g., 768 dimensions~\cite{sbert, clip}) require aggressive quantization to fit into memory, which reduces recall and necessitates a second-pass refinement that retrieves full-precision vectors from SSD~\cite{fusionann, cagra_note}. Figure~\ref{fig:bg_prof} shows this refinement dominates latency on GPU-accelerated pipeline~\cite{cuvs} where the index and quantized vectors reside in GPU memory while full-precision vectors are mmaped on the host. On an A10 GPU with a 40-thread CPU, index traversal accounts for only 2–15\% of query time due to GPU acceleration, whereas refinement dominates due to random SSD I/O and distance computation. If all vectors could reside in main memory, an impractical assumption at large scale, performance would improve by up to $14\times$. This unattainable upper bound motivates our progressive distance estimation approach, which incrementally refines candidates using compact residual codes: reducing full-vector fetches, cutting refinement overhead, and avoiding the capacity ceilings of hardware-centric solutions.

\subsection{Vector Quantization and Distance Estimation}
Vector quantization compresses high-dimensional vectors into compact codes, reducing storage cost and enabling efficient distance computation. The simplest form, \textit{scalar quantization} (SQ), discretizes each dimension independently, but scales poorly with vector length. \textit{Product quantization} (PQ)~\cite{adc,opq,faiss} improves scalability by partitioning a vector into subspaces and by quantizing each with a separate codebook, enabling efficient table-based distance lookups and GPU acceleration~\cite{faiss,cagra,fusionann,bang}. \textit{Residual quantization} (RQ)~\cite{irq,trq} further enhances accuracy by encoding a vector as a coarse approximation plus successive residual corrections, often using PQ at each stage.  
To estimate distances, most systems employ \textit{asymmetric distance computation} (ADC)~\cite{pq}, which compares a full-precision query with quantized database codes:
\[
\hat{d}(q, x) = d\!\left(q,\sum\nolimits_{i=1}^L Q^{-1}(c_i)\right),
\]
where $c_i$ is the i-th layer RQ code. While effective, existing systems~\cite{faiss, cuvs} decode all quantization levels for every vector during traversal, even though most candidates are filtered out shortly thereafter. As a result, 10-100$\times$ more candidates are decoded than necessary, wasting bandwidth on vectors that are immediately discarded. \design addresses this inefficiency by enabling progressive ADC over residual quantization: early RQ levels are stored in fast memory while deeper residual codes are tiered into the slower far memory. Distance estimation proceeds incrementally, allowing early pruning of unpromising candidates and avoiding full reconstruction for most vectors.

\subsection{Related Works}

Recent work reduces ANNS memory cost by leveraging high-bandwidth memory systems such as CXL~\cite{cxlanns}, persistent memory~\cite{hmann}, near-data-processing (NDP)~\cite{nmp1,nmp2,nmp3, nmp4}, and GPUs~\cite{fusionann,bang,cagra}. They either accelerate table lookups on quantized codes or assume full-precision vectors are directly available, leaving refinement to slow storage reads. CXL-ANNS~\cite{cxlanns} stores full vectors in a CXL memory and offloads distance computation to FPGA, but its scalability collapses once the original dataset cannot fit the memory capacity. HM-ANN~\cite{hmann} maps HNSW~\cite{hnsw} index entry layers to fast DRAM–NVM tiers, but for high-dimensional vectors, additional layers provide little pruning benefit while increasing management overhead~\cite{hiearchy_hub}. 
NDP designs~\cite{nmp1,nmp2,nmp3, nmp4} integrate distance computation with quantized vectors directly into memory devices, but leave refinement unaddressed. NDP accelerations are complementary to our work.  We focus on progressive, tier-aware refinement that minimizes SSD access and can naturally benefit from or be combined with NDP acceleration across fast and far memory.

\section{\design: Estimate the distance progressively}

\begin{figure}[t]
    \centering
    \includegraphics[width=0.95\linewidth]{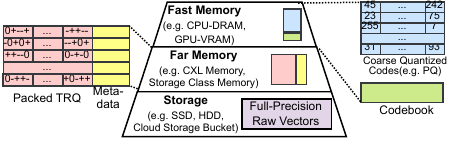}
    \vspace{-12pt}
    \caption{Memory-tiered layout of the \design framework}
    \vspace{-18pt}
    \label{fig:format}
\end{figure}

To reduce the costly I/O and computation of second-pass refinement, we propose \design, a tier-aware framework that incrementally sharpens coarse distance estimates. As shown in Figure~\ref{fig:format}, coarse PQ codes and the codebook remain in fast memory, while compact residual codes and metadata are stored in far memory, reducing high-latency SSD accesses. The refinement builds on a decomposition of the L2 distance into coarse and residual terms, which can be progressively estimated without vector reconstruction. Residuals are encoded into compact ternary codes that eliminate multiplications and pack efficiently into far memory. This compact format enables progressive refinement: first by residual distance estimation, then by multiplication-free encoding, and finally by an enhanced estimator that integrates all components into accurate L2 distance. Together, these steps allow \design to prune non-promising candidates early and cut SSD traffic.

\subsection{L2 Distance Decomposition}
\label{sec:l2_dist_decomp}
Unlike prior systems~\cite{diskann,cagra_note,bang} that perform reranking using a separate quantization code built solely for refinement, \design progressively reuses the coarse approximation already computed in earlier stages. This reuse avoids discarding useful information from the coarse code and enables refinement to be expressed cleanly through an L2 distance decomposition:
\[\|x - q\|_2^2 = \|x_c - q\|_2^2 + \|x_c - x\|_2^2 - 2\langle q - x_c, x - x_c \rangle,\]
where $q$ is the query, $x_c$ is the reconstructed vector of original vector $x$ from the coarse code, and $\ip{\cdot}{\cdot}$ is the inner product. The distance is split into three terms: the coarse approximation, the compression distortion, and a residual inner product.

\begin{figure}[t]
    \centering
    \includegraphics[width=0.8\linewidth]{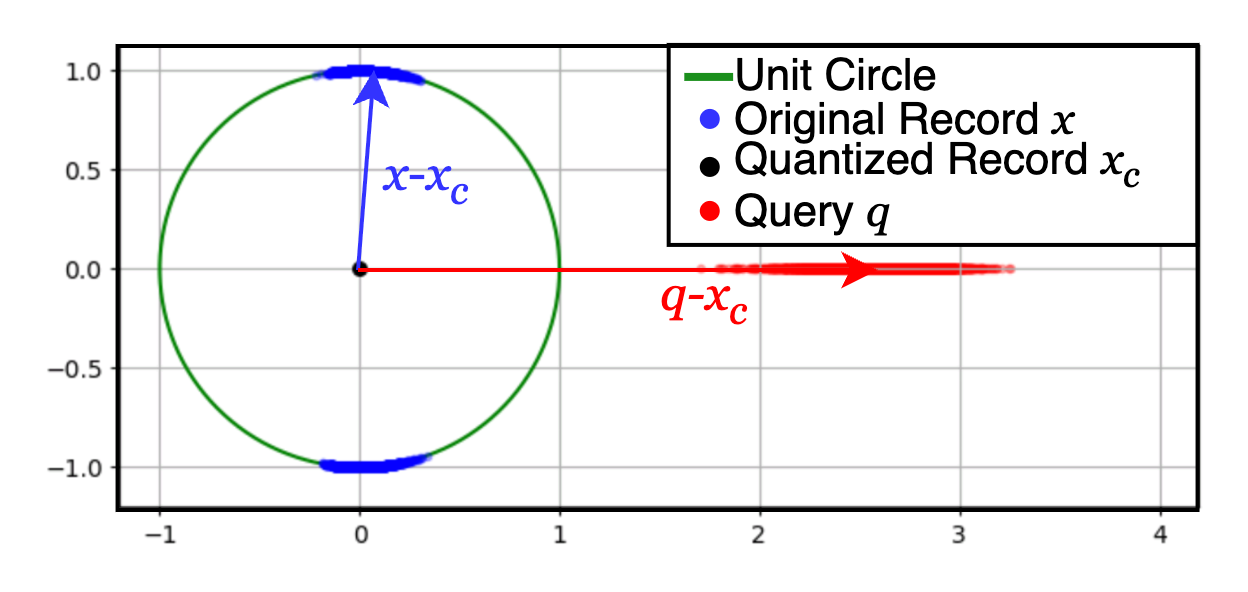}
    \vspace{-16pt}
    \caption{Visualization of the residual vector and query vector}
    \label{fig:level1_quantization}
    \vspace{-12pt}
\end{figure}

Figure~\ref{fig:level1_quantization} shows this decomposition on the Wiki dataset~\cite{wiki}. 
For each record vector, we align its reconstruction of coarse quantization $x_c$ to the origin (circle center) and scale the residuals $x - x_c$ to lie on the unit circle. The query offset $q - x_c$ is placed along the x-axis, with length proportional to  $\frac{\|q - x_c\|}{\|x - x_c\|}$. 
In this representation, the original records $x$ (blue points) lie on the unit circle, 
and queries $q$ (red points) are located along the x-axis. It highlights that the residual is nearly orthogonal to the query offset, so their inner product is small.

This leads to a \textit{first-order approximation}:
\[\hat{d}_1(q,x) = \|x_c - q\|_2^2 + \|x_c - x\|_2^2 = \hat{d}_0(q,x) + \|x_c - x\|_2^2,\]

where $\hat{d_0}(q,x)=\|x_c - q\|_2^2$. Note that $\|x_c - x\|_2^2$ is a scalar value associated with each record vector and can be precomputed offline. To fully recover the true distance, we must estimate residual item $\ip{q-x_c}{x-x_c}$.

Letting $\delta = x-x_c$ denote the residual between the record and its coarse approximation, we have 
$$\lTwoSq{x-q} = \lTwoSq{q-x_c} + \lTwoSq{\delta} + 2\ip{x_c}{\delta} - 2\ip{q}{\delta}.$$

The first three terms can be computed using only the coarse quantization code and precomputed scalars. The last term, $\ip{q}{\delta}$, is estimated via quantized residuals without decoding the base vector. Compared to storing full raw vectors, residuals are smaller, since the base code already captures most of the vector structure. Storing them in far memory offers a better trade-off between latency and capacity. We refer to this refinement using quantized residual vectors as the \textit{second-order approximation}. 
Since residual quantization is naturally stackable, distance estimates can be progressively refined. For example, we can first encode the residual on top of the coarse code, and then refine it further by encoding finer residuals on the remaining error, enabling progressively tighter distance estimates.
Without loss of generality, we focus on second-order estimation in the following discussion. The next section details how $\delta$ is quantized and used to obtain an unbiased estimate of the inner product term.

\subsection{\design Residual Distance Estimation}

To estimate the inner product $\ip{q}{\delta}$, we first quantize the residual vector $\delta$ to a codeword $\delta_c$, and have
\[
\ip{q}{\delta}
= \lTwo{q}\,\lTwo{\delta}\,\ip{e_q}{e_\delta},
\]
where $e_q = q/\lTwo{q}$ and $e_\delta = \delta/\lTwo{\delta}$ are the normalized directions of the query q  and the residual $\delta$.  Our goal is then to obtain an unbiased estimator of the directional term $\ip{e_q}{e_\delta}$.

Observe that by adding and subtracting the projection of $e_q$ onto $e_{\delta_c}$, the direction of $\delta_c$, we can write
\begin{align}
\langle e_q, e_\delta \rangle 
&= \langle e_q, e_\delta \rangle + \langle e_q, e_{\delta_c} \rangle \langle e_\delta, e_{\delta_c} \rangle - \langle e_q, e_{\delta_c} \rangle \langle e_\delta, e_{\delta_c} \rangle \nonumber \\
&= \langle e_q, e_{\delta_c} \rangle \langle e_{\delta_c}, e_\delta \rangle + \langle e_q - \langle e_q, e_{\delta_c} \rangle e_{\delta_c}, e_\delta \rangle  \nonumber \\
&=  \ip{e_q}{e_{\delta_c}}\ip{e_{\delta_c}}{e_\delta} + \lTwo{ e_q - \ip{e_q}{e_{\delta_c}}} \ip{e_\perp}{e_\delta}  .
\label{eq:s}
\end{align}

This equation decomposes $e_q$ into components aligned with and orthogonal to $e_{\delta_c}$. $e_\perp$ is the unit vector orthogonal to $e_{\delta_c}$ in the plane spanned by $e_{\delta_c}$ and $e_q$.

This can be viewed as the dual form of the query disaggregation of previous work~\cite{rabitq} that disaggregates it to the original direction $e_{\delta}$, instead of the quantized  direction $e_{\delta_c}$ in our work, and corresponding orthogonal basis. That work shows the orthogonal term (second term of Equation \ref{eq:s}) is concentrated and its expectation is zero under mild conditions, specifically, when residual directions are evenly distributed and uncorrelated with the query. In our case, because the coarse quantization already captures most of the vector similarity, the residual vectors $\delta$ are nearly isotropic and uncorrelated with the query. This removes the need for an additional random projection step, allowing us to exploit the information encoded in each residual code well. Consequently, we estimate the first term while treating the second as an error term having zero expectation, yielding an unbiased residual distance estimator.

\subsection{\design Multiplication-Free Encoding}

To enable efficient estimation of $\ip{q}{\delta}$, \design encodes the residual direction with a sparse ternary vector. This compact representation can avoid multiplications during refinement and reduce the memory footprint. Concretely, given a residual vector $e_\delta\in\mathbb{R}^D$, we seek the code $\bar e_{\delta_c}$ in the codebase $\mathcal{C}=\{-1,0,1\}^D$, so that its normalized version 
\[
e_{\delta_c} \;=\;\frac{\bar e_{\delta_c}}{\lTwo{\bar e_{\delta_c}}}
\]
minimizes $\|e_\delta - e_{\delta_c}\|_2$.  Observing that 
\[
\bigl\|e_\delta - e_{\delta_c}\bigr\|_2^2
=2 - 2\,\ip{e_{\delta_c}}{e_\delta}
=2 - 2\,\ip{\tfrac{\bar e_{\delta_c}}{\lTwo{\bar e_{\delta_c}}}}{e_\delta},
\]
we equivalently choose
\[
\bar e_{\delta_c}
=\arg\max_{c\in\mathcal{C}}
\ip{\frac{c}{\lTwo{c}}}{e_\delta}
=\arg\max_{c\in\mathcal{C}}
\frac{\sum_{i=1}^D c_i\,e_{\delta,i}}{\lTwo{c}}.
\]
Direct enumeration of all $3^D$ candidates is infeasible. We solve this optimization by observing that any optimal $c$ must select exactly $k$ nonzero entries (each set to the sign of $e_{\delta,i}$) among the $k$ largest magnitudes of $e_\delta$. It would strictly decrease the numerator of the inner product if including any smaller-magnitude $e_{\delta,i}$ in place of a larger one, while changing the total count of nonzero entries $k$ only affects the denominator $\sqrt{k}$.  Thus, the best choice for a given $k$ is to take the first $k$ entries of the sorted list. Concretely, by sorting the absolute values into 
$x_1\ge x_2\ge \cdots\ge x_D,$ where $ x_i =\bigl|e_{\delta,(i)}\bigr|$, we reduce the inner‐product term to 
\[
\ip{\frac{c}{\lTwo{c}}}{x}
=\frac{\sum_{i=1}^k x_i}{\sqrt{k}}
\quad\text{whenever }\sum_i c_i^2 = k.
\]
Therefore, the global maximization over $\mathcal{C}$ collapses to  one‐dimensional search
\[
\max_{c\in\mathcal{C}}
\ip{\frac{c}{\lTwo{c}}}{x}
\;=\;
\max_{1\le k\le D}\;\frac{\sum_{i=1}^k x_i}{\sqrt{k}}.
\]
The proposed approach requires only $O(D\log D)$ time to sort $x_i$ values, $O(D)$ time to build the prefix sums of $x_i$,
and another $O(D)$ pass to evaluate each ratio $S_k/\sqrt{k}$ and identify the maximizer $k^*$.  Once $k^*$ is known, we form $\bar e_{\delta_c}$ by assigning $\operatorname{sign}(e_{\delta,(i)})$ to the top $k^*$ entries and zeros elsewhere, then normalize by $\sqrt{k^*}$.  This procedure yields the exact optimal ternary codeword in $O(D\log D)$ time without ever enumerating the full codebook $\mathcal{C}$.

\subsection{Compact \design Code}

Figure~\ref{fig:format} illustrates the data layout used by \design to store per-record information in far memory during the refinement phase. Each record includes two scalar values: $\ip{x_c}{\delta}$ and $\lTwo{\delta}$. They serve as scaling metadata for computing the final distance estimate, along with one quantized residual vector $\bar e_{\delta_c}$ that captures the remaining error after level-1 approximation.

The residual vector $\bar e_{\delta_c}$ contains elements from a 3-level quantization set $\{-1, 0, 1\}$. Each element, therefore, requires more than 1 bit to encode, but it is unnecessary and wasteful to use a full 2 bits per value, as that would encode 4 states while only 3 are needed. The entropy of a uniformly distributed 3-level variable is $\log_2 3 \approx 1.58$ bits, which sets the theoretical limit for lossless compression.

To approach this bound in practice, \design uses a fixed-length compact encoding where every 5 dimensions of the residual vector are packed into a single byte. This is achieved by first mapping each value $x_i \in \{-1, 0, 1\}$ to the set $\{0, 1, 2\}$ using the transformation $x_i + 1$, and then encoding the resulting 5-digit base-3 number into a byte using the formula below. This encoding yields a per-dimension storage cost of 1.6 bits.
\[
y = \sum_{i=0}^{4} 3^i (x_i + 1).
\]

\subsection{\design Enhanced Refinement Distance Estimator}
\label{sec:de}

With the decomposition of the distance function and the residual encoding already introduced, the final step is to wrap these components into the refinement estimator. Specifically, the L2 distance can be expressed as a combination of the coarse approximation $\hat{d_0}$, precomputed scalar $\lTwoSq{\delta}$ and $2\ip{x_c}{\delta}$, and the estimated residual term $-2\ip{q}{\delta}$.

An important insight emerges when evaluating the effectiveness of these estimations. Lower mean squared error (MSE) in distance approximation does not always lead to higher ANNS recall. This is because recall is  primarily determined by how accurately candidates are ranked near the top-$k$ decision boundary. Large estimation errors for very close or very distant points have little impact as they do not affect ranking within the critical region. Thus, improving global distance accuracy does not guarantee better recall; what matters is local precision around the decision boundary.

Based on this observation, \design learns a lightweight linear calibration model offline that directly optimizes for recall in the boundary region by subsampling a small calibration set of sample–neighbor pairs.  Specifically, it randomly draws a small number of database vectors(about 0.3\% in practice, which we found sufficient to saturate the model).
For each sampled vector $x$, it leverages the existing index to identify approximate neighbors without requiring a costly exact kNN search: if the index is graph-based (e.g., CAGRA~\cite{cagra} or HNSW~\cite{hnsw}), it uses the graph-adjacent vertices of $x$; if the index is IVF-based, it uses the vectors in the same inverted list. These points naturally cluster near $x$ in the original space and provide dense coverage of its local decision boundary.

For each sample–neighbor pair, we construct feature vector:
\[
A = \bigl[\;\hat d_0,\;\hat d_{\mathit{ip}},\;\lTwo{\delta}^2,\;\langle x_c,\delta \rangle\;\bigr],
\]
where $\hat d_0 = \|q-x_c\|_2^2$ is the coarse approximation, $\hat d_{\mathit{ip}}$ is the refined estimate of $-2\langle q,\delta \rangle$ via ternary coding, $\lTwo{\delta }^2$ is the residual norm, $\langle x_c,\delta \rangle$ is the precomputed cross‐term.
Let $D$ denote the ground-truth distance.  The calibration model learns in the offline stage by solving
\[
\hat W \;=\;\arg\min_W \bigl\|D - A\,W\bigr\|_2^2,
\]
via ordinary least squares. At query time, refinement distance estimation reduces to the lightweight computation of $A \hat W$.

\begin{figure}[t]
    \centering
    \includegraphics[width=0.95\linewidth]{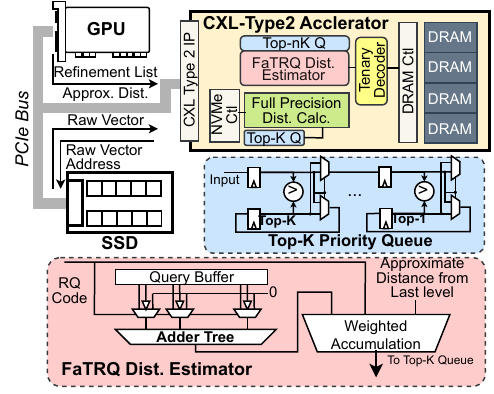}
    \vspace{-6pt}
    \caption{System architecture of the \design-augmented ANNS pipeline.}
    \vspace{-12pt}
    \label{fig:arch}
\end{figure}

\section{\design-augmented ANNS system}

While \design targets a broad range of tiered memory systems, we also explore its benefits on modern far-memory devices. A CXL Type-2 module offers both the large capacity of far-memory (e.g. SCM) and, additionally, lightweight computation. Our prototype integrates customized refinement logic into such a device, pushing refinement directly into far memory to reduce host–device data movement. As compute-enabled large memory devices have already been commercialized~\cite{cxl_device}, this demonstrates how \design can exploit the emerging hardware model, while our software implementation shows it is not tied to specialized far-memory hardware.

As shown in Figure~\ref{fig:arch}, the front-stage GPU processes the index using coarse quantization codes to estimate approximate distances and generates a large candidate list. Instead of transferring hundreds of bytes of coarse code, only 4-byte coarse distance values per candidate are sent to the far-memory device equipped with the \design refinement logic. Unlike prior approaches~\cite{irq,trq} that reconstruct record vectors from both coarse and residual codes, \design refines directly on compact residual codes, avoiding reconstruction. The far-memory accelerator reranks the candidates using its on-device distance estimator and trims the list based on updated scores. Only the top results are then fetched from SSD for final computation, cutting refinement cost by 85\% (Section~\ref{sec:res_overall}).

Figure~\ref{fig:arch} also shows the detailed implementation. Two hardware priority queues, implemented using registers and comparators, are used to track the top-$K$ nearest neighbors during refinement. 
One queue maintains the top‑$K$ candidates ranked by estimated distances from \design's residual quantization path, while the other is used for final ranking based on full-precision vector distances. Each entry in the queue stores a distance value and a pointer to the corresponding vector. New candidates are inserted by comparing their distance to those in the queue, and bubbling smaller values forward through the pipeline of comparators. 
Each queue supports up to 1024 entries. A 256-entry lookup table implements the ternary decoder, which unpacks residual codes into their ternary representation. Because the residual codes in \design are ternary, the inner product $\ip{e_q}{e_{\delta_c}}$ between the query and residual vector can be computed using simple adders and multiplexers. A weighted accumulation unit, implemented as a MAC array, combines these results to produce the final estimated distance, as described in Section~\ref{sec:de}.

\begin{figure*}[ht]
    \centering
    \begin{minipage}[t]{0.38\textwidth}
        \centering
        \includegraphics[width=\linewidth]{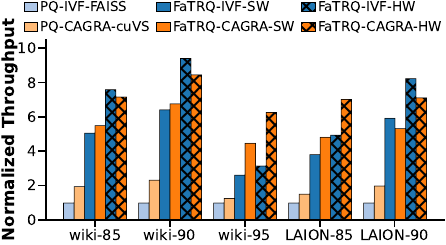}
        \vspace{-18pt}
        \caption{\design normalized throughput evaluation for different top-10 query recall rates on both IVF and CAGRA front-stage indexes.}
        \vspace{-15pt}

        \label{fig:throughput}
    \end{minipage}\hfill
    \begin{minipage}[t]{0.27\textwidth}
        \centering
        \includegraphics[width=\linewidth]{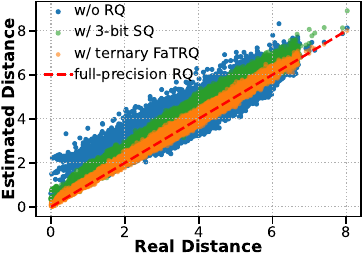}
        \vspace{-18pt}
        \caption{The distortion in squared Euclidean distance estimation relative to the top-100 ground truth results.}
        \vspace{-15pt}

        \label{fig:sq_trq}
    \end{minipage}\hfill
    \begin{minipage}[t]{0.27\textwidth}
        \centering
        \includegraphics[width=\linewidth]{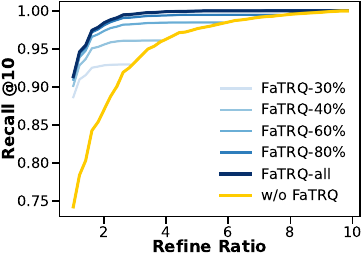}
        \vspace{-18pt}
        \caption{Recall@10 versus refinement ratio under varying proportions of candidates filtered with \design.}
        \vspace{-15pt}

        \label{fig:refine_vs_reall}
    \end{minipage}
\end{figure*}

\section{Evaluation}
\subsection{Experimental Setup}

\begin{table}[t]
\centering
\small
\caption{Parameters for \design Evaluation}
\vspace{-4pt}
\label{tab:eval}
\begin{tabular}{|l|c|}
\hline
\textbf{Parameter}                & \textbf{Value}     \\ \hline
DRAM Configuration                & 8Gb x16 DDR5-4800  \\ \hline
Timing (tRCD-tCAS-tRP)            & 34-34-34           \\ \hline
Channels / Ranks per Channel      & 8 / 8              \\ \hline
SSD Latency / Throughput ~\cite{ssd_parm} & 45 $\si{\mu s}$ / 1200K IOPS \\ \hline
CXL Latency / Throughput ~\cite{cxl_device}  & 271 $\si{n s}$ : 22 $\si{GB/s}$ \\ \hline
\end{tabular}
\vspace{-12pt}
\end{table}

\textbf{Software and Baselines:} We implement \design as an extension to the cuVS~\cite{cuvs} and FAISS~\cite{faiss} libraries. Our design stacks the proposed residual codes on top of product quantization (PQ) as a coarse quantizer, and integrates them with both IVF~\cite{faiss} and CAGRA~\cite{cagra} index structures to form the complete ANNS system. For evaluation, we use the FAISS GPU release for IVF and the GPU-oriented graph index CAGRA from cuVS. All parameters are tuned via grid search~\cite{cagra}. We compare these \design-enhanced systems against the baseline pipelines provided in cuVS~\cite{cagra, cagra_note}.

\textbf{Datasets:} We use two pre-embedded ANNS datasets. a) The \textit{Wiki} dataset~\cite{wiki} (251~GB) contains 88M SBERT~\cite{sbert} multilingual sentence embeddings (768-D). b) The \textit{LAION} dataset~\cite{laion}, the largest pre-embedded collection from VDBBench~\cite{vdbench}, contains 100M CLIP~\cite{clip} embeddings derived from LAION-5B~\cite{laion}, totaling 286~GB. Both datasets use Euclidean distance as the similarity metric with 10k queries.

\textbf{Platform:}  We extend Ramulator~\cite{ramulator} to simulate far-memory access patterns of a CXL Type-2 device, with DRAM parameters listed in Table~\ref{tab:eval}. Hardware overhead is assessed by synthesizing our accelerator in Verilog at 1 GHz using the ASAP7~\cite{ASAP7}. On-chip SRAM is modeled with FinCACTI~\cite{fincacti}.
All GPU-side index traversal and coarse distance computations run on an NVIDIA A10 GPU (24 GB VRAM), while refinement in the baseline systems executes on a 40-thread Intel Xeon Gold 6230 CPU with 128 GB DRAM and a 1 TB SSD.

\subsection{Overall Performance}
\label{sec:res_overall}

Figure~\ref{fig:throughput} compares end-to-end throughput of \design against SoTA GPU pipelines, IVF-FAISS~\cite{faiss} and CAGRA-cuVS~\cite{cuvs}, at 85\%, 90\%, and 95\% recall for top-10 queries. On the LAION, recall saturates at 94\% when PQ codes fit in GPU VRAM, so results for LAION-95 are omitted. The baselines store quantized vectors in GPU memory and perform refinement on the CPU by fetching full-precision vectors from SSDs. We evaluate \design with residual codes stored in far memory: in the software mode (-SW), codes reside in CXL memory, but filtering is done on the CPU, while in the hardware mode (-HW), filtering is offloaded to a CXL Type-2 accelerator.

\design-HW delivers $3.1\times$–$9.4\times$ speedup over IVF-FAISS and $2.6\times$–$4.9\times$ over CAGRA-cuVS. The benefit is larger with IVF because it requires more refinements. For example, at 90\% recall on Wiki, IVF refines 320 candidates per query, compared to 120 for CAGRA. In the baseline, this means 320 vs. 120 SSD fetches. With \design, those become 28 SSD plus 320 CXL accesses for IVF, and 17 SSD plus 120 CXL accesses for CAGRA, shifting most refinement traffic off SSD and yielding bigger improvements for IVF. The speedup narrows at 95\% recall, as deeper traversal and coarse filtering required for high accuracy begin to dominate runtime. We also observe the hardware variant adds $1.2\times$–$1.5\times$ higher throughput over the software path, with candidate filtering up to $3.7\times$ faster due to direct far-memory access and removal of host data movement.

\subsection{Evaluation on Distance Estimation Distortion}

As discussed in Section~\ref{sec:de}, the accuracy of ANNS depends on the distance estimation between the query and a small subset of the collection that is close to the query. To evaluate the distance distortion introduced by \design, we calculate the distance between each query and the top-100 ground truth results obtained via exhaustive search. Figure~\ref{fig:sq_trq} compares distance distortion on the Wiki dataset for three cases: INT8 quantization(w/o RQ), PQ with scalar quantization (SQ) residual codes~\cite{bang}, and PQ with \design residual codes. Both SQ and \design reuse the same PQ codebook from the earlier experiment, ensuring consistency across methods.

Unlike SQ-based residuals, which reconstruct record vectors, \design refines distances directly via enhanced estimation without transferring upper-level codes. As shown, \design stays much closer to the oracle line (full-precision residual vectors) than 3-bit SQ, with an MSE of $0.0159$ vs. $0.258$.

Moreover, \design achieves better storage efficiency. For a 768-dimensional vector, \design requires only $768/5 + 8 = 162$ bytes (packing five ternary values into a byte, plus four bytes for precomputed values), compared to $768 \times 4 / 8 = 384$ bytes needed by 4-bit SQ for comparable MSE ($0.0134$).

\subsection{Refinement Reduction Analysis}

We studied how the filtering ratio impacts candidate reranking. For each query, we collected the true top-100 based on PQ distances and examined reranking behavior. 
Figure~\ref{fig:refine_vs_reall} plots recall against the refinement ratio, defined as the SSD IO normalized by the final top-$k=10$. 
Each blue curve corresponds to a filtering rate, where only the top-$X$\% of the FaTRQ-ranked queue accesses full-precision vectors; the yellow curve is the baseline where the entire list requires SSD I/O. 
Without \design, recovering the true top-10 with 99\% probability requires scanning up to 70 full-precision vectors out of 100 candidates.  With \design, the same guarantee is reached within 25, reducing the refinement by $2.8\times$.

\subsection{Overhead Analysis}

The customized processing unit integrated into the CXL Type-2 device adds only 0.729 $\si{mm^2}$ area and 897 $\si{mW}$ power. The \design distance estimator accounts for 29\% of the area and 27\% of the power, while the priority queue adds 6\% and 8\%. Compared to a CXL memory controller~\cite{cxl_ctrl} with 16 ARM Neoverse V2 cores (2.5 $\si{mm^2}$, 1.4 $\si{W}$ each~\cite{arm_area}), the overhead is under 1.8\% in area and 4\% in power, showing \design a lightweight addition to the whole memory expander.

The offline overhead of \design is also minimal. Training the lightweight calibration model and constructing residual codes requires only a single parallel pass per vector, adding about 10 minutes in total, compared to roughly 3 hours to build the CAGRA index for either dataset in our experiments.

\section{Conclusion}


We introduce \design, a tier-aware refinement layer that compresses residuals into ternary codes and incrementally refines coarse distances without reconstructing full vectors. \design targets a key bottleneck in modern ANNS systems: second-pass refinement that fetches full-precision vectors from slow storage. As embedding dimensionality and dataset scale continue to grow, this I/O cost increasingly dominates query latency. By streaming compact residuals from far memory and updating distances on the fly, \design enables early candidate pruning and cuts unnecessary fetches, boosting throughput by up to $9\times$ compared to GPU\cite{cagra}.

\section*{Acknowledgment}
This work was supported in part by PRISM and CoCoSys—centers in JUMP 2.0, an SRC program sponsored by DARPA; by the U.S. DOE DeCoDe Project No. 84245 at PNNL; and by the NSF under Grants No. 2112665, 2112167, 2052809, and 2211386.


\bibliographystyle{IEEEtran} 
\bibliography{IEEEabrv, references}    

\end{document}